\documentclass[conference]{IEEEtran}
\IEEEoverridecommandlockouts
\usepackage{cite}
\usepackage{amsmath,amssymb,amsfonts}
\usepackage{algorithmic}
\usepackage{graphicx}
\usepackage{textcomp}
\usepackage{xcolor}
\usepackage{makecell}
\usepackage{bm}
\usepackage{dsfont}

\def\BibTeX{{\rm B\kern-.05em{\sc i\kern-.025em b}\kern-.08em
    T\kern-.1667em\lower.7ex\hbox{E}\kern-.125emX}}

\begin{document}

\title{Masked Privileged-Information Distillation for Multimodal Skin Lesion Classification Under Missing Clinical Metadata\\
}

\author{\IEEEauthorblockN{Anirban Barua, Md Mahir Abrar Khan, Ayman Iktidar, Md. Sajjatul Islam}
\IEEEauthorblockA{\textit{
Department of Computer Science and Engineering} \\
\textit{Chittagong Independent University (CIU), Chattogram, Bangladesh} \\
\{anirban\_barua, mahir\_abrar\_khan, ayman\_iktidar, sajjatulislam\}@ciu.edu.bd}
}

\maketitle

\begin{abstract}

Multimodal skin lesion classification combines clinical images with patient metadata to improve diagnostic accuracy. However, complete metadata available during training may be only partially accessible at deployment, and resource-constrained settings additionally require computational efficiency. We address these challenges with a privileged-information distillation framework in which a multimodal teacher trained on complete metadata supervises a 9.2$\times$ smaller student trained with randomly masked clinical fields. Clinical fields are masked as whole groups at a per-sample rate drawn from U(0,1), so one training run covers the full metadata availability range. Fusion is residual, with metadata added as a gated correction to an unconditional image base. On the PAD-UFES-20 dataset, distillation under masked training improves balanced accuracy over cross-entropy training at every availability level, by an average of +4.7 points versus +1.8 points without masking. The masked student loses only 7.8 balanced-accuracy points as metadata decreases from complete to absent, compared with 36.6 points for the same student trained on complete metadata, highlighting the role of masked training in graceful degradation beyond distillation alone. Grad-CAM visualizations further show that the masked student's attention generally remains lesion-centered as metadata is withdrawn. The resulting compact model targets point-of-care settings, where clinical metadata is often incomplete.

\end{abstract}


\section{Introduction}
Skin cancer is one of the most commonly diagnosed cancers worldwide, with over 337,000 new melanoma cases and over 1.1 million non-melanoma cases globally~\cite{sung2026globocan}, emphasizing the need for efficient classification methods to facilitate early diagnosis. Yet, the number of lesions requiring assessment far exceeds what specialist services can evaluate directly, especially in regions where patients are first seen by general practitioners or community health workers rather than dermatologists~\cite{giavinabianchi2020}. Automated classification from clinical images offers a way to extend triage capacity into these settings. However, these settings often have limited infrastructure and thus require models to run on commodity hardware or mobile devices at the point of care. This creates a fundamental constraint on model size and inference cost, regardless of any accuracy gains that might be obtained by using larger models.

Multimodal classifiers improve on image-only models by incorporating patient metadata, such as age, lesion location, diameter, and other clinically relevant information that provides diagnostic cues beyond visual appearance. This reflects clinical practice, where physicians combine visual examination with patient history rather than relying on images alone. In deployment, however, not all metadata used to develop a machine learning model is present. Patients may be unable to provide details such as family history, intake forms may be incomplete, and certain fields may be omitted due to time constraints. Thus, a model trained and optimized under the assumption of full metadata may perform poorly when deployed in these variable conditions. This, together with the computational constraint discussed above, motivates the development of a single lightweight model that can robustly handle varying levels of metadata availability without the need for retraining.

Robustness to missing metadata and model compactness have largely been treated as separate concerns in multimodal skin lesion classification. Rather than treating the development to deployment gap as a limitation, we exploit complete training-time metadata as privileged information to guide a student model designed to operate across varying levels of metadata availability at inference. Using the PAD-UFES-20 dataset~\cite{padufes20}, this work trains a
multimodal ConvNeXt-Base teacher on complete metadata and distills it
into a smaller and faster ConvNeXt-Pico student that sees only a
randomly masked subset of clinical fields, enabling it to achieve both
a small footprint and tolerance for incomplete records. Realistic
missingness is simulated by randomly varying the metadata availability rate
from 0\% to 100\%, allowing a single training procedure to handle the
full range of metadata availability. Fusion is residual, so absent metadata can only fail to contribute, never suppress the image signal. The result is a single compact model that degrades gracefully as metadata is withheld, rather than collapsing as conventional multimodal classifiers do under the same conditions. 

Our contributions are:
\begin{itemize}

    \item We design a distillation framework that treats a teacher's complete metadata during training as privileged information for the student, masking clinical fields at a random availability rate $r \sim \mathcal{U}(0,1)$ so that a single model spans the full range of metadata availability without retraining. It uses residual gated fusion, where the image provides the base representation and metadata contributes a gated correction.

\item We show that masked training combined with distillation drives robustness: masked training yields far more graceful degradation than unmasked training as metadata is withdrawn, and distillation adds a substantially larger accuracy gain under masking than without it.

    \item We deliver a distilled ConvNeXt-Pico student: $9.2\times$ smaller and $6.2\times$ faster on CPU than the teacher, and more robust than the teacher at low metadata availability.
\end{itemize}

\section{Related work}

Prior work on multimodal skin lesion classification on PAD-UFES-20 used various strategies for fusing clinical images with patient metadata. One of the earliest works, MetaBlock~\cite{metablock2021} introduced an LSTM-gate-inspired attention module that scales and suppresses CNN feature maps using one-hot-encoded metadata, yielding a gain of at least 8.2 percentage points in balanced accuracy (BACC) over image-only baselines. It was validated across five CNN backbones, namely EfficientNet-B4, DenseNet-121, MobileNet-V2, ResNet-50, and VGG-13. 

MetaBlock-SE~\cite{metablockse2025} replaced one-hot encoding with SBERT sentence embeddings of free-text metadata, reaching 70.2\% BACC on PAD-UFES-20 while remaining robust across simulated missingness rates of 0--70\%, though performance still fell below image-only baselines beyond 50\% dropout. It is the only prior work to address metadata missingness directly, but does so by training and evaluating a separate model at each fixed availability rate, rather than a single model spanning the full continuum at deployment.

AtnMetaBlock~\cite{atnmetablock2025} applied Lasso-filtered clinical vectors within a cross-attention module, reporting a maximum 10.4-point BACC gain over baselines and reaching 78.5\% BACC with a VGG-13 backbone, while MiSC~\cite{misc2025} adopted a late-fusion design concatenating the class probabilities of separately trained image and metadata classifiers, transferring reasonably to HAM10000 and ISIC-2019. Both assume complete metadata at inference and do not report patient-level splitting.

DualRefNet~\cite{dualrefnet2025} applied a dual-stage refinement strategy: an auxiliary super-resolution task for the visual encoder and class-frequency-based weight regularization of the final classifier layers, achieving strong performance. As with the methods above, it requires metadata at test time and does not report patient-level splitting, leaving open the risk of patient leakage between training and test partitions. More recent studies focus on efficiency and reliability: TriMNet~\cite{trimnet2026} replaced attention-based metadata encoders with a Mamba-based state-space encoder, reaching 82.7\% classification accuracy at near-linear complexity, and DermaCalibra~\cite{dermacalibra2026} combined Bayesian uncertainty estimation with
dynamic feature modulation and integrated-gradient attribution, using a 23.6M-parameter model. DermaCalibra addresses missing fields only through simple imputation, without a missingness ablation to validate its claimed graceful degradation.

None of the methods discussed above treat missing metadata as more than a preprocessing or robustness concern; missingness is handled by architectural choice, but never formalized within a principled teacher-student framework. Knowledge distillation (KD)~\cite{hinton2015} transfers the soft output distribution of a large teacher network into a smaller student, improving the student's generalization beyond what its capacity alone would allow. Independently, the learning-using-privileged-information (LUPI) paradigm considers a teacher with access to additional per-instance information available only at training time. Lopez-Paz et al.~\cite{lopezpaz2016} unified these two views as generalized distillation, showing that KD and LUPI are special cases of a single teacher-student framework in which a teacher trained on privileged features $x^*$ transfers knowledge to a student restricted to the original input $x$. This formulation applies directly to clinical metadata, which is naturally privileged: available during model development but often missing or unreliable at deployment.

We build on generalized distillation by treating clinical metadata as privileged information across the full range of availability, rather than assuming it is available at a fixed rate. To our knowledge, no
prior work on this task trains a single distilled student that covers the full
range of deployment-time metadata availability without retraining.

\section{Methodology}

\begin{figure*}[htbp]
    \centering
    \includegraphics[width=0.9\textwidth]{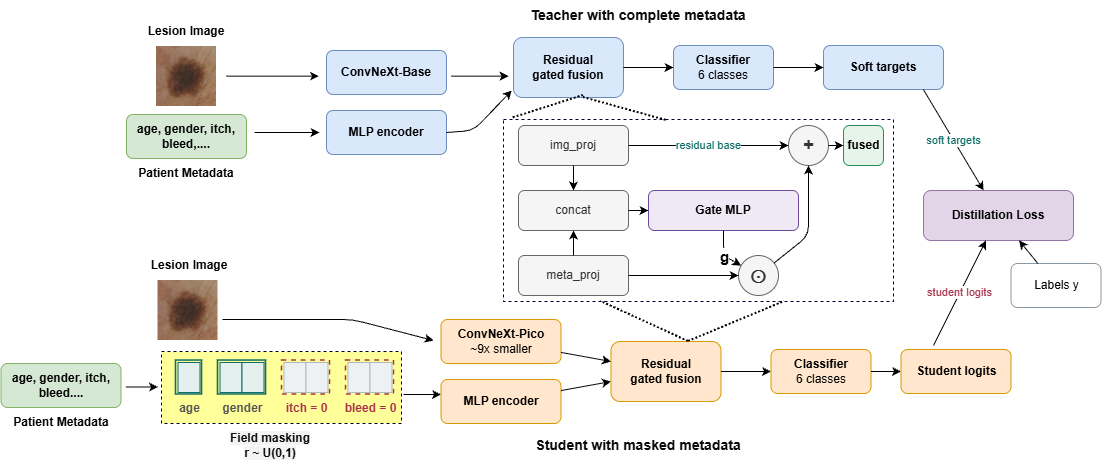}
    \caption{Overview of the proposed privileged-information distillation framework}
    \label{fig}
\end{figure*}

\subsection{Dataset}\label{AA}
Our method is evaluated on PAD-UFES-20, a dataset of 2298 smartphone-acquired clinical images containing six diagnostic classes: actinic keratosis (ACK), basal cell carcinoma (BCC), melanoma (MEL), nevus (NEV), squamous cell carcinoma (SCC), and seborrheic keratosis (SEK). Each image is matched with a clinical questionnaire of 21 fields, including age, anatomical region, lesion diameter, symptom indicators, and personal and family history. The dataset is very imbalanced, with the two most frequent classes accounting for the majority of images (Table~\ref{tab1}). This imbalance is addressed via class-weighted training and balanced-accuracy evaluation throughout. A stratified 70/15/15 patient-grouped train/validation/test split is used, based on each patient's modal diagnosis, ensuring that all images from a single patient are entirely contained within a single partition, avoiding data leakage and providing a more difficult and realistic evaluation than the image-level protocols often used in prior work on this dataset.

\begin{table}[htbp]
\caption{Data distribution in train, validation, and test sets}
\begin{center}
\begin{tabular}{c c c c}
\hline
\textbf{Class} & \textbf{Train} & \textbf{Validation} & \textbf{Test}\\
\hline
ACK & 523 & 106 & 101 \\
BCC & 590 & 126 & 129 \\
MEL & 38 & 6 & 8 \\
NEV & 168 & 37 & 39 \\
SCC & 131 & 31 & 30 \\
SEK & 158 & 41 & 36 \\
\hline
\textbf{Total} & \textbf{1608} & \textbf{347} & \textbf{343} \\
\hline
\end{tabular}
\label{tab1}
\end{center}
\end{table}

\subsection{Preprocessing}
All images are resized to 256$\times$256, cropped to 224$\times$224, and normalised using ImageNet statistics. Training images are augmented with random horizontal/vertical flips, rotations up to 15°, and colour jitter, while validation and test images receive no augmentation. The 21 clinical fields comprise three continuous attributes (age and two lesion diameters) and 18 categorical attributes. Continuous fields are standardized to zero mean and unit variance using training-split statistics only. Categorical fields are one-hot encoded into contiguous field-specific blocks, producing an 81-dimensional metadata vector. Following the dataset’s encoding convention, unanswered entries are dropped rather than represented as explicit categories, resulting in all-zero blocks. Our masking scheme reproduces this pattern, making masked categorical fields indistinguishable from genuinely unanswered ones.

\subsection{Architecture}
\label{sec:arch}

Teacher and student share a similar architecture, differing only in image
backbone: a ConvNeXt-Base~\cite{convnext} encoder (CLIP-LAION2B pretrained)
for the teacher and a roughly nine-times-smaller, ConvNeXt-Pico encoder for the student. An input image is encoded to
$\mathbf{f}_{\text{img}}$ by the backbone, and the $81$-dimensional metadata
vector is encoded by a three-layer MLP ($81\!\to\!512\!\to\!256\!\to\!256$,
with batch normalisation, ReLU, and dropout $p{=}0.1$ after each hidden layer)
to $\mathbf{f}_{\text{meta}}$.

\textbf{Residual Gated Fusion:} The two modalities are combined so that the
image is an unconditional base and metadata enters only as a gated correction.
Both features are first projected to a common $512$-dimensional space,
$\mathbf{h}_{\text{img}} = W_{\text{img}}\mathbf{f}_{\text{img}}$ and
$\mathbf{h}_{\text{meta}} = W_{\text{meta}}\mathbf{f}_{\text{meta}}$. A gate
$\mathbf{g}$ is then computed from their concatenation by a two-layer MLP and
used to scale the metadata projection, which is added to the image projection (Eqs.~1--2):
\begin{align}
\mathbf{g} &= \sigma\!\left(W_2\,\phi\!\left(W_1[\mathbf{h}_{\text{img}}; \mathbf{h}_{\text{meta}}]\right)\right) \in \bm{\mathbb{R}^{512}} \label{eq:gate}\\
\mathbf{h}_{\text{fused}} &= \mathbf{h}_{\text{img}} + \mathbf{g} \odot \mathbf{h}_{\text{meta}} \in \bm{\mathbb{R}^{512}} \label{eq:fused}
\end{align}
where $\phi$ is ReLU, $\sigma$ the logistic sigmoid, and $\odot$ the
element-wise product. The gate can reach two extremes:
$\mathbf{g} = \mathbf{0}$ yields a purely image-based fused representation,
while $\mathbf{g} = \mathbf{1}$ admits the full metadata contribution.
The fused vector $\mathbf{h}_{\text{fused}}$ is classified by a two-layer head
($512\!\to\!256\!\to\!6$, with dropout $p{=}0.3$ before the first layer, ReLU,
and dropout $p{=}0.2$ before the second) into the six diagnostic classes.

Because $\mathbf{h}_{\text{img}}$ is never scaled by the gate, the metadata
branch can only add to the fused representation. Under missing metadata the gated correction may become
uninformative, but the image representation reaches the classifier intact. We
also compare against a concatenation baseline that projects
$[\mathbf{h}_{\text{img}}; \mathbf{h}_{\text{meta}}]$ jointly through a linear
layer followed by ReLU, weighting both modalities unconditionally.

\subsection{Field-Group Masking}
We train the student to handle varying levels of metadata completeness by masking whole clinical fields rather than individual encoded columns.
The $81$ columns are partitioned into $21$ groups, one per clinical field. Let
$\mathbf{m} \in \mathbb{R}^{81}$ be the encoded metadata and
$g(j) \in \{1,\dots,21\}$ map column $j$ to its field. We draw a
per-sample availability rate $r \sim \mathcal{U}(0,1)$ and one uniform variate
per field, keeping a field when its variate falls below $r$ (Eqs.~3--4):
\begin{align}
k_j &= \mathds{1}\left[u_{g(j)} < r\right], \quad u_i \overset{\text{iid}}{\sim} \mathcal{U}(0,1) \\
\tilde{\mathbf{m}} &= \mathbf{m} \odot \mathbf{k} \in \bm{\mathbb{R}^{81}} \label{eq:mask}
\end{align}
Since $k_j$ depends only on $u_{g(j)}$, every column of a field is kept or zeroed together, resulting in the all-zero pattern the encoding reserves for an unanswered question. Here $r$ is the expected
fraction of fields present, so one training run spans the full range from all fields present ($r{=}1$) to none ($r{=}0$), with masks redrawn each batch. Field-level masking covers the full range of metadata completeness while keeping the inputs valid. Modality dropout only works at the endpoints,
$r \in \{0,1\}$, whereas column-level dropout can mask one column of a field while leaving another active, producing an encoding that no real record could have.

\subsection{Privileged-Information Distillation}
\label{sec:distill}

The teacher is trained once on complete metadata and then kept frozen. During
distillation, it always receives the unmasked record and produces soft targets, while the student sees only the masked metadata. This difference in what the two models can access is what enables privileged information: the teacher can use information that is unavailable to the student, helping the student learn to make reliable predictions even when some metadata is missing. 

For a training image with label $y$, let $\mathbf{z}^{t}$ and $\mathbf{z}^{s}$
be the teacher and student logits, the teacher computed from complete metadata
and the student from the masked vector $\tilde{\mathbf{m}}$. The student is
trained with a convex combination of a supervised term and a distillation term (Eqs.~5--6):
\begin{align}
\mathcal{L} &= (1-\alpha)\,\mathcal{L}_{\text{CE}}
+ \alpha\,T^{2}\,\mathrm{KL}\!\left(p^{t}_{T} \,\|\, p^{s}_{T}\right) \\
p^{t}_{T} &= \mathrm{softmax}(\mathbf{z}^{t}/T),
\qquad
p^{s}_{T} = \mathrm{softmax}(\mathbf{z}^{s}/T)
\end{align}
where $\mathcal{L}_{\text{CE}}$ is class-weighted cross-entropy on the student
logits, $T$ is the distillation temperature, and $\alpha$ balances the two
terms. Class weights are assigned inversely to their frequency in the training set and normalised so that they sum to the number of classes. This gives rarer classes, such as melanoma, greater influence during training. The same weighting scheme is used across all arms, ensuring that it does not affect the comparison between them.

We train two arms: the \emph{distilled} arm uses $\alpha=0.5$, learning from both the labels and the teacher's soft targets, while the \emph{cross-entropy} arm uses $\alpha=0$, learning from the labels alone under the same masking.
 The mask of equation
\eqref{eq:mask} is seeded only by epoch, batch index, and batch size,
independent of the run seed and of which model is training, so at a fixed
seed the two arms receive identical masks every batch. Because the masks are
paired across arms and every other factor is held fixed, the difference
between the two isolates the effect of privileged supervision.

To determine whether robustness comes from masked training itself or from distillation, we additionally train an unmasked variant of each arm, each receiving complete metadata throughout training ($r = 1$). Evaluation still sweeps the full availability range for every model. Both arms are checkpointed on a mixed-availability
validation score, the mean balanced accuracy over $r \in \{0, 0.5, 1\}$, rather
than on full-metadata validation. Selecting on full metadata would favour
checkpoints that rely heavily on complete records, which is the opposite of our target deployment setting. Instead, the mixed criterion rewards checkpoints that perform well across the full availability range.

\section{Results and Discussion}

\subsection{Experimental Settings}
All models are trained on a dual-NVIDIA T4 GPU setup. Both teacher and students
use the AdamW optimiser with a base learning rate of $10^{-4}$, weight decay
$10^{-4}$, and a batch size of $32$, with the pretrained backbone frozen for the
first five epochs and a lower learning rate applied to backbone parameters
thereafter. The learning rate follows a cosine schedule after unfreezing. Models are trained for up to $40$ epochs, with early
stopping on the validation criterion. Distillation uses temperature $T = 2$ and
balance $\alpha = 0.5$. Each configuration is run over five seeds; we report mean and standard deviation across them. Balanced accuracy (BACC) is our primary metric throughout, given the dataset's class imbalance.

\subsection{Fusion Ablation}
\label{sec:fusion-ablation}

Before distillation, we select the teacher architecture by comparing fusion
strategies at full metadata availability, all using the ConvNeXt-Base backbone
and trained under identical settings. Table~\ref{tab:fusion} reports balanced
accuracy and macro AUC.

\begin{table}[htbp]
\caption{Performance of ConvNeXt-Base models at full metadata availability}
\begin{center}
\begin{tabular}{c|c|c}
\hline
\textbf{Model} & \textbf{\textit{BACC}} & \textbf{\textit{Macro AUC}} \\
\hline
Metadata only & 0.628 $\pm$ 0.039 & 0.895 $\pm$ 0.007 \\
Image only & 0.673 $\pm$ 0.036 & 0.931 $\pm$ 0.003 \\
MetaBlock~\cite{metablock2021} & 0.758 $\pm$ 0.030 & 0.948 $\pm$ 0.008 \\
Concatenation & 0.790 $\pm$ 0.033 & 0.963 $\pm$ 0.005 \\
\textbf{Residual Gated Fusion} & \textbf{0.821 $\pm$ 0.006} & \textbf{0.965 $\pm$ 0.001} \\
\hline
\end{tabular}
\label{tab:fusion}
\end{center}
\end{table}

We observe that either modality alone is limited: metadata-only and image-only reach $0.628$ and $0.673$ BACC, respectively. All three fusion variants improve on image-only, though the gain depends heavily on the mechanism. MetaBlock, adapted to pooled features since a direct implementation of the original performed poorly, reaches only $0.758$, below concatenation ($0.790$). Our proposed residual gated fusion reaches $0.821$, a $3.1$-point gain over concatenation and $6.3$ points over adapted MetaBlock, and is markedly more stable across seeds, supporting it as the teacher fusion mechanism.

\subsection{Degradation Across Metadata Availability}
\label{sec:results}

\begin{figure*}[!t]
\centerline{\includegraphics[width=.8\textwidth]{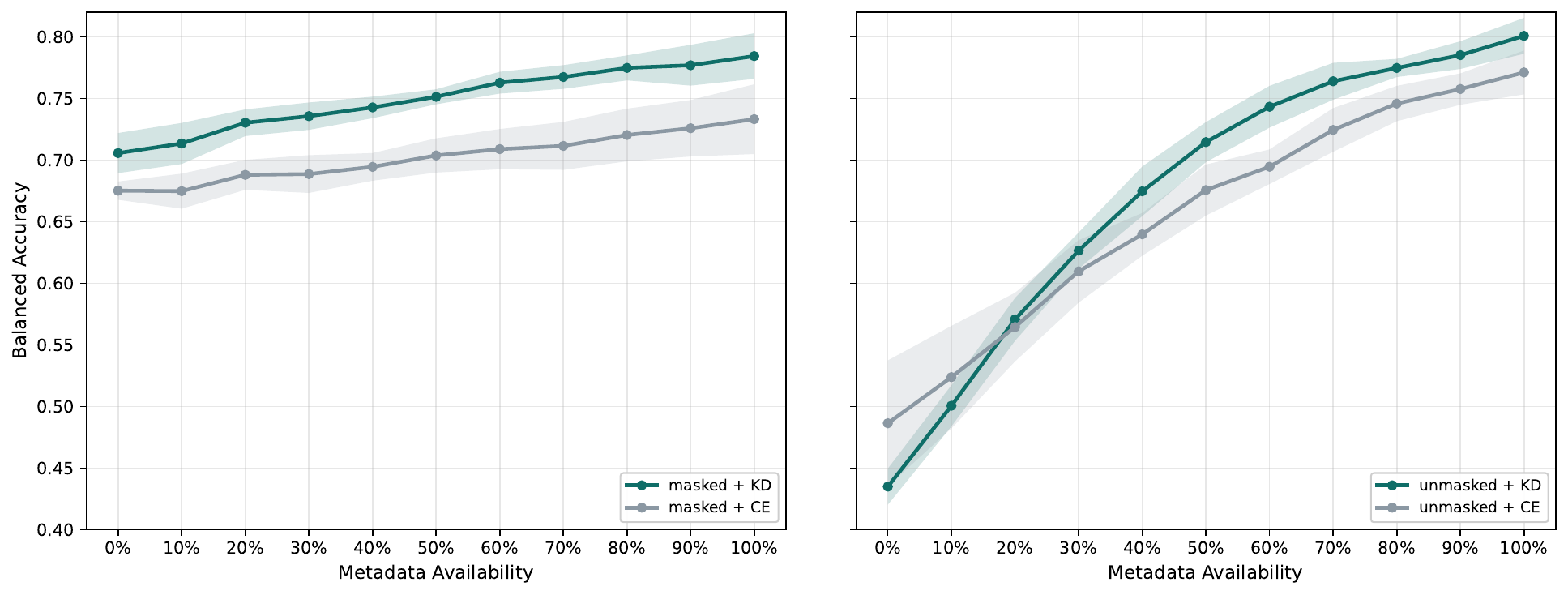}}
\caption{Balanced accuracy across metadata availability for distilled and
cross-entropy students, under masked (left) and unmasked (right) training}
\label{fig:sweep}
\end{figure*}

We evaluate every model along a metadata-availability sweep, $r \in \{0.0,
0.1, 0.2, \ldots, 1.0\}$ (11 levels). For each target rate, fields are
independently kept with probability $r$; the resulting balanced accuracy is
averaged over five mask draws, with the deterministic endpoints $r \in
\{0, 1\}$ evaluated once. Fig.~\ref{fig:sweep} plots the full
sweep; Tables~\ref{tab:masked} and~\ref{tab:unmasked} report only the even
levels for space, though all statistics quoted in the text are computed
over the complete sweep.

\begin{table}[htbp]
\caption{Balanced accuracy of the distilled (KD) and cross-entropy (CE)
students under masked training}
\begin{center}
\begin{tabular}{c|c|c}
\hline
\textbf{\textit{Availability}} & \textbf{\textit{KD (Proposed)}} & \textbf{\textit{CE}} \\
\hline
0.0 & \textbf{0.706 $\pm$ 0.016} & 0.675 $\pm$ 0.007 \\
0.2 & \textbf{0.730 $\pm$ 0.011} & 0.688 $\pm$ 0.012 \\
0.4 & \textbf{0.743 $\pm$ 0.009} & 0.695 $\pm$ 0.011 \\
0.6 & \textbf{0.763 $\pm$ 0.009} & 0.709 $\pm$ 0.016 \\
0.8 & \textbf{0.775 $\pm$ 0.010} & 0.720 $\pm$ 0.021 \\
1.0 & \textbf{0.784 $\pm$ 0.019} & 0.733 $\pm$ 0.028 \\
\hline
\end{tabular}
\label{tab:masked}
\end{center}
\end{table}
It is evident that the distilled student is more accurate at every availability level over CE arm in the masked protocol, for an average gain of
$4.7$ points across the full sweep of 11 levels. Additionally, for both KD and CE arms, accuracy declines monotonically as
metadata is withdrawn, yet neither collapses. The KD advantage is smallest at zero availability and grows as more metadata becomes available, showing the teacher's supervision helps most when the student has substantial metadata to contextualize it. Notably, the masked KD student at zero metadata availability exceeds even the teacher's own image-only ablation (0.673 BACC), despite using a much smaller backbone.
\begin{table}[htbp]
\caption{Balanced accuracy of the distilled (KD) and cross-entropy (CE)
students under unmasked training}
\begin{center}
\begin{tabular}{c|c|c}
\hline
\textbf{\textit{Availability}} & \textbf{\textit{KD (Proposed)}} & \textbf{\textit{CE}} \\
\hline
0.0 & 0.435 $\pm$ 0.015 & \textbf{0.487 $\pm$ 0.051} \\
0.2 & \textbf{0.571 $\pm$ 0.017} & 0.565 $\pm$ 0.028 \\
0.4 & \textbf{0.675 $\pm$ 0.020} & 0.640 $\pm$ 0.017 \\
0.6 & \textbf{0.743 $\pm$ 0.017} & 0.695 $\pm$ 0.014 \\
0.8 & \textbf{0.775 $\pm$ 0.007} & 0.746 $\pm$ 0.014 \\
1.0 & \textbf{0.801 $\pm$ 0.015} & 0.771 $\pm$ 0.018 \\
\hline
\end{tabular}
\label{tab:unmasked}
\end{center}
\end{table}

To confirm that the robustness comes from masked
training and not from distillation alone, we train both arms without masking, on
complete metadata, and evaluate them along the same sweep
levels. At high availability the unmasked models match or
slightly exceed their masked counterparts, as expected when the test-time input
resembles training. Also, the gap between KD and CE arms is much lower with an average of only $1.8$ points. As availability falls they collapse: unmasked KD drops from
$0.801$ at full metadata to $0.435$ at zero, a $36.6$-point span, against a
$7.8$-point span for the masked student. At zero availability the masked student
outperforms the unmasked one by $27.1$ points. Masked training, not distillation, is therefore what produces graceful degradation; distillation improves accuracy on top of it.

\subsection{Model Compactness}
\label{sec:compactness}

Table~\ref{tab:compact} compares the teacher and the distilled student on size, compute, and accuracy. The teacher is a single frozen checkpoint, the lowest of five seeds (five seed BACC mean $0.821$); student values are means over five seeds.

\begin{table}[htbp]
\caption{Teacher and distilled student on size, compute, and accuracy.}
\begin{center}
\begin{tabular}{c|c|c}
\hline
\textbf{\textit{Metric}} & \textbf{\textit{Teacher}} & \textbf{\textit{Student (Proposed)}} \\
\hline
Parameters (M) & 89.0 & \textbf{9.7} \\
Size (MB) & 356.2 & \textbf{38.9} \\
GMACs & 15.4 & \textbf{1.4} \\
CPU latency (ms) & 217.8 & \textbf{35.4} \\
GPU latency (ms) & 13.6 & \textbf{5.2} \\
BACC ($r{=}0$) & 0.452 & \textbf{0.706} \\
BACC ($r{=}0.5$) & 0.679 & \textbf{0.751} \\
BACC ($r{=}1$) & \textbf{0.814} & 0.784 \\
\hline
\end{tabular}
\label{tab:compact}
\end{center}
\end{table}

The student is $9.2\times$ smaller in parameters and on disk, needs $11\times$ fewer multiply-accumulate operations, and runs $6.2\times$ faster on CPU, the setting that matters for point-of-care deployment on commodity hardware. At full metadata the student trails the teacher by $3.0$ points, of which $1.3$ reflect model capacity (an unmasked student reaches $0.801$) and $1.7$ the cost of training for missing metadata. In return, the student is far more robust when metadata is withheld: at $r{=}0$ it retains $0.706$ BACC, while the teacher falls to $0.452$, a $25.4$-point difference, narrowing to $7.2$ points at $r{=}0.5$.

\subsection{Qualitative Analysis}
\label{sec:qualitative}
\begin{figure}[htbp]
    \centering
    \includegraphics[width=0.9\columnwidth]{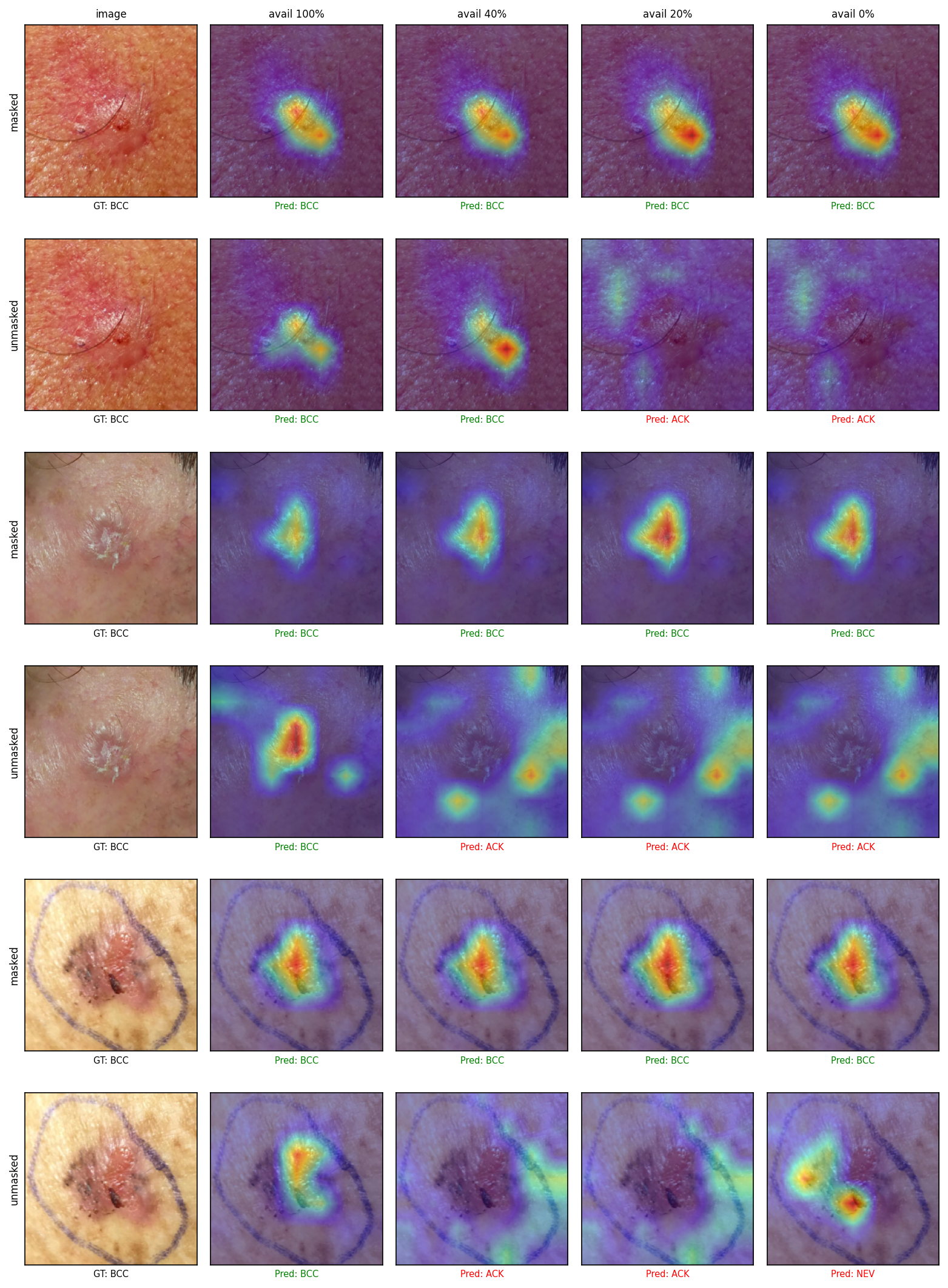}
    \caption{Grad-CAM attributions for the masked and unmasked students across metadata availability, for three lesions.}
    \label{fig:gradcam}
\end{figure}

We compare Grad-CAM~\cite{gradcam} attributions for the masked and unmasked KD
students on the same lesions across the availability sweep to understand why masked training degrades more gracefully than unmasked training. This pattern is common and consistent across
training seeds: averaged over five seeds, $96.6 \pm 9.1$ ($\sim$28\% of test set) lesions are classified
correctly by the masked student at every availability level but
misclassified by the unmasked student at zero. Fig.~\ref{fig:gradcam} shows
three representative cases. The masked student's attribution remains
lesion-centered throughout the sweep, at times growing sharper as metadata is
withdrawn, consistent with the model compensating for missing metadata by
relying more heavily on visual evidence. In contrast, the unmasked student's attribution increasingly shifts toward features unrelated to the lesion as metadata becomes less available, including background hair and a surgical marker spot. These shifts occur alongside prediction errors, consistent with the performance collapse for unmasked training.

\section{Conclusion \& Future Work}
We addressed the mismatch between complete metadata available during
training and partial metadata available at deployment, together with the
computational constraints of real world skin lesion diagnosis, via a
privileged-information distillation framework in which a ConvNeXt-Base
teacher trained on complete metadata supervises a compact ConvNeXt-Pico
student trained under randomly masked metadata,
fused through a residual gated formulation. On PAD-UFES-20, masked training was the main factor enabling graceful degradation as metadata became unavailable, while distillation provided an additional performance benefit. In contrast, the same student architecture trained without masking was substantially less robust to missing metadata. Despite its smaller size and lower computational cost, the resulting student remained competitive with the teacher and performed better when metadata was withheld. Grad-CAM analysis further indicated that the masked student's attention remained more consistently centred on the lesion as metadata became unavailable, supporting the observed robustness. Future work includes external validation on independent clinical datasets and exploration of more efficient architectures for resource-constrained clinical deployment.

\vspace{12pt}

\end{document}